\documentclass[conference]{IEEEtran}
\IEEEoverridecommandlockouts

\usepackage{cite}
\usepackage{amsmath,amssymb,amsfonts}
\usepackage{algorithmic}
\usepackage{graphicx}
\usepackage{textcomp}
\usepackage{xcolor}
\def\BibTeX{{\rm B\kern-.05em{\sc i\kern-.025em b}\kern-.08em
    T\kern-.1667em\lower.7ex\hbox{E}\kern-.125emX}}
\begin{document}

\title{Learning Dark Souls Combat Through Pixel Input With Neuroevolution\\}

\author{\IEEEauthorblockN{Jim O'Connor}
\IEEEauthorblockA{\textit{Computer Science Department} \\
\textit{Connecticut College}\\
New London, Connecticut \\
joconno2@conncoll.edu}
\and
\IEEEauthorblockN{Gary B. Parker}
\IEEEauthorblockA{\textit{Computer Science Department} \\
\textit{Connecticut College}\\
New London, Connecticut \\
parker@conncoll.edu}
\and
\IEEEauthorblockN{Mustafa Bugti}
\IEEEauthorblockA{\textit{Computer Science Department} \\
\textit{Connecticut College}\\
New London, Connecticut \\
mbugti@conncoll.edu}
\and
}


\maketitle

\begin{abstract}
This paper investigates the application of Neuroevolution of Augmenting Topologies (NEAT) to automate gameplay in Dark Souls, a notoriously challenging action role-playing game characterized by complex combat mechanics, dynamic environments, and high-dimensional visual inputs. Unlike traditional reinforcement learning or game playing approaches, our method evolves neural networks directly from raw pixel data, circumventing the need for explicit game-state information. To facilitate this approach, we introduce the Dark Souls API (DSAPI), a novel Python framework leveraging real-time computer vision techniques for extracting critical game metrics, including player and enemy health states. Using NEAT, agents evolve effective combat strategies for defeating the Asylum Demon, the game's initial boss, without predefined behaviors or domain-specific heuristics. Experimental results demonstrate that evolved agents achieve up to a 35\% success rate, indicating the viability of neuroevolution in addressing complex, visually intricate gameplay scenarios. This work represents an interesting application of vision-based neuroevolution, highlighting its potential use in a wide range of challenging game environments lacking direct API support or well-defined state representations.
\end{abstract}


\section{Introduction}

The development of artificial intelligence (AI) capable of playing video games at a human or superhuman level has long been an important benchmark in AI research \cite{risi2015neuroevolution, miikkulainen2006computational}. From early rule-based systems to modern deep learning approaches, various techniques have been used to train autonomous agents capable of solving complex game environments. Among these, neuroevolution \cite{stanley2019designing} is an approach that evolves artificial neural networks using evolutionary algorithms, making it a promising method, particularly in cases where reward structures are sparse or environmental dynamics are not easily modeled \cite{floreano2008neuroevolution}.

Traditional AI approaches to game playing have been successfully applied to well-defined rule-based environments such as chess \cite{campbell2002deep}, Go \cite{silver2017mastering}, and Atari games \cite{schrittwieser2020mastering}, where state representations are either explicitly provided or easily inferred. However, more complex modern video games, particularly those lacking official APIs, present significant challenges for AI-driven gameplay \cite{pearce2022counter}. These games require agents to interpret high-dimensional visual information, navigate complex and nonlinear world structures, and respond dynamically to unpredictable enemy behaviors. Traditional reinforcement learning methods like Deep Q-Learning (DQN) \cite{mnih2015human} often struggle when rewards are sparse or delayed, as is common in challenging 3D games like Dark Souls, due to difficulties in propagating meaningful error signals through large convolutional neural networks. In contrast, NEAT addresses these limitations by directly optimizing network structure and weights through evolutionary selection, thereby bypassing the credit assignment problem inherent in RL and potentially leading to more rapid adaptation to complex visual inputs. This capability makes NEAT particularly promising for games where direct reward signals are limited or non-existent. \cite{such2017deep, salimans2017evolution}.

This paper investigates the application of Neuroevolution of Augmenting Topologies (NEAT) \cite{stanley2002evolving} to a notoriously difficult video game, Dark Souls \cite{darksoulsremastered}. Unlike many other games used in AI research, Dark Souls does not provide easily accessible state representations, requiring agents to learn directly from raw pixel data. Additionally, the difficulty of Dark Souls is a fundamental part of its design philosophy\cite{clark2024dark}, built upon the legacy of classic action RPGs and influenced by game mechanics that emphasize deliberate combat and meticulous level design. Overcoming these mechanics through the creation of AI agents presents a unique challenge for evaluating the capabilities of neuroevolutionary methods in high-dimensional, visually complex environments.

The methodology presented in this work builds upon previous studies in evolutionary computation and neuroevolution. Early implementations of evolutionary algorithms in game-playing AI, such as Fogel’s work on evolving neural networks for board games \cite{chellapilla1999evolution}, demonstrated the potential of artificial evolution in decision-making tasks. More recent applications, including the HyperNEAT Atari General Game Player \cite{hausknecht2012hyperneat}, showcased the viability of neuroevolution for large-scale pixel-based game environments. However, much of this prior work has focused on relatively constrained two-dimensional games with well-defined objectives. By extending these principles to Dark Souls, we seek to further push the catalogue of games approached by neuroevolution techniques and concepts. 

In this paper, we detail the development of a NEAT-based agent trained exclusively through visual input to defeat the first boss of Dark Souls. The agent begins with no prior knowledge of the game world, learning through evolutionary selection based on in-game performance metrics.
\section{Methodology}

Released in 2011 by FromSoftware, Dark Souls is a third-person action role-playing game renowned for its punishing difficulty, intricate level design, and deep combat mechanics \cite{clark2024dark}. It has achieved notable acclaim for its challenging gameplay, which requires players to develop precise timing, pattern recognition, and adaptive strategies to overcome formidable enemies and environmental hazards.

To address these challenges, we developed the Dark Souls API (DSAPI)\footnote{Available at: https://github.com/ConnAALL/darksoulsapi}, a Python-based interface designed to facilitate AI interaction with the game. DSAPI enables direct extraction of key game metrics from the user interface, such as player health, enemy health, and the in-game currency known as Souls. It also provides functionality for navigating menus and initializing the game at specific checkpoints, allowing for more efficient training and evaluation of AI agents. By leveraging computer vision techniques, DSAPI processes real-time screen capture to extract necessary gameplay data, eliminating the need for manual intervention or external game modifications.

Through this framework, we aim to demonstrate the feasibility of applying neuroevolutionary methods, and more specifically NEAT, to Dark Souls, paving the way for further research into AI-driven gameplay in highly complex and visually intricate gaming environments.

\subsection{Neuroevolution of Augmenting Topologies}
NEAT is an evolutionary algorithm designed to optimize artificial neural networks by evolving both their structure and weights. Unlike traditional fixed-topology neural networks, NEAT begins with minimal architectures and incrementally increases their complexity over generations. This is achieved through mutation operations that add new nodes and connections, allowing for adaptive growth in response to environmental challenges. Speciation ensures that effective structures are preserved by grouping similar networks together, and a crossover mechanism allows genetic material to be recombined, fostering diversity in the population.

NEAT leverages historical markings to track genes through generations, ensuring effective crossover between compatible structures. It also dynamically adjusts mutation rates, allowing networks to evolve complexity in a controlled manner. This is particularly useful in environments where optimal network structure is not known ahead of time. In Dark Souls, where the agent must infer spatial relationships and control dynamics from raw pixel input, NEAT provides a powerful mechanism for discovering optimal actions through iterative refinement.

\begin{figure}
    \centering
    \includegraphics[width=1\linewidth]{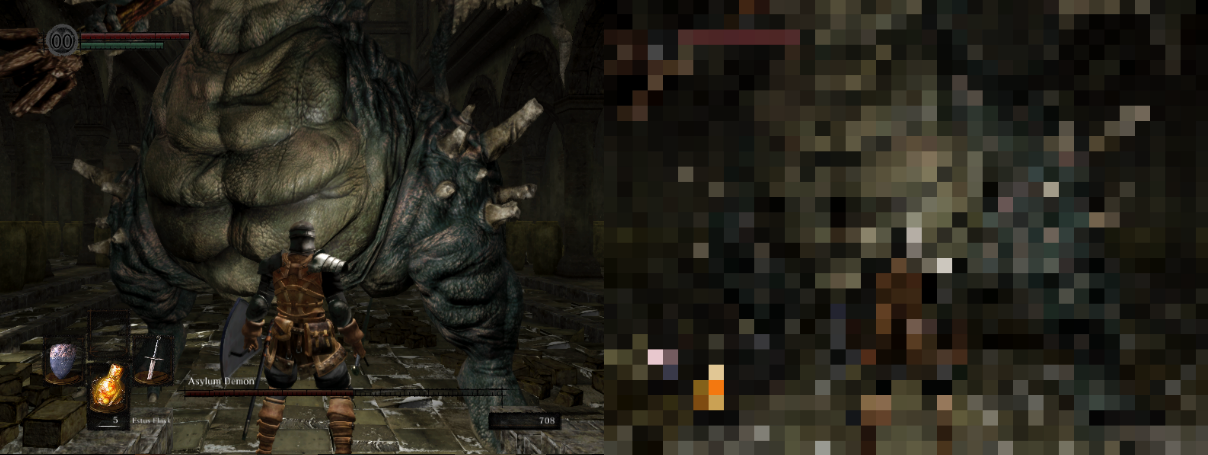}
    \caption{The processed image serves as input for the neural network. The downscaling of the screenshot greatly reduces input size while retaining enough important features to keep the game playable by a human player. This approach improves computational efficiency and allows even low-end computers to run and train in the game.}
    \label{fig:colorinput}
\end{figure}

To apply NEAT to Dark Souls, we designed a neural network with raw visual input processed through simple downscaling, as shown in Figure 1. The network outputs a discrete set of control actions, corresponding to movement, attacks, and defensive maneuvers. Fitness is evaluated based on damage dealt and damage taken. Over generations, evolved networks develop sophisticated strategies, demonstrating the potential of NEAT for complex real-time decision-making.

\section{Learning Environment and Experimental Design}
Dark Souls presents significant challenges for AI research due to the absence of official interfaces or accessible programmatic control. To overcome this issue, we developed the previously mentioned DSAPI, a Python framework that enables interaction with Dark Souls Remastered through real-time image processing and screen capture. DSAPI relies on standard computer vision techniques, such as template matching and pixel-level analysis, to detect and monitor essential in-game information, including player health, enemy health, menu navigation, and game state transitions.

Prior to each evaluation, the environment is reset to an identical initial condition to ensure consistency across agent trials, enabling reliable fitness comparisons. The reset procedure consists of exiting to the game's main menu, restoring a predefined game save via direct manipulation of the game save directory, and navigating game menus using automated keyboard inputs determined by visual recognition of menu elements. 

The Asylum Demon, selected as the initial scenario, is the introductory boss of Dark Souls and presents a combat scenario that requires fundamental gameplay strategies, including timed attacks and evasive actions. This scenario simplifies the broader complexities of Dark Souls by generally excluding camera control and environmental navigation, thereby isolating combat strategy as the primary learning objective.

\subsection{Experimental Procedure and Game Loop}
The evaluation process for each neuroevolutionary trial follows a structured game loop designed to facilitate controlled agent interaction with Dark Souls. Each trial commences with an environment reset as previously described.

The neural network generates a probability distribution over a discrete action set consisting of fourteen actions. These actions are movements and rolling maneuvers in the cardinal directions, backstepping, standard and heavy attacks, shield blocking and parrying, and healing with the Estus Flask.

DSAPI translates the highest-probability action output from the neural network into corresponding keyboard inputs executed within the game. These inputs are executed at the rate that the specified algorithm is able to process the pixel input and output an action. For our specified NEAT network, this translates to roughly 5 actions a second. It is important to note that these actions may not all be followed through, as in-game actions take a fixed duration to complete, regardless of subsequent decisions made by the AI. For example, a strong attack takes a much longer time to complete than a parry action.

As such, the polling frequency of the agent is also based on the efficiency of the algorithm being used to interface with DSAPI. While DSAPI can provide pixel input at a high frame rate, the efficiency of one's algorithm will determine how many frames can be utilized. In a real-time scenario, this encourages the use of efficient algorithms for processing input and executing actions.

Subsequently, fitness evaluation is based on player and boss health information extracted through computer vision methods applied to the respective health bars, as shown in Figure 2. The fitness function used in these experiments is defined as follows:

\begin{equation}
Fitness = P + (100 - B)
\end{equation}

Where $P$ is equal to the player's percentage of total health, and $B$ is equal to the boss's percentage of total health, both out of 100. The function rewards survival (player health) and offensive combat effectiveness (boss damage), with agents receiving zero fitness if they die without inflicting damage. This formulation explicitly prevents the emergence of passive or non-engaging behaviors. This evaluation is only carried out once the terminal state is reached, and agents do not have access to their fitness.
Each trial continues until a terminal state is reached, defined by player or boss defeat. These events are identified through template matching against in-game notifications ("YOU DIED" or "VICTORY ACHIEVED" popups). Upon termination, the simulation environment resets in preparation for the next evaluation.

\begin{figure}
    \centering
    \includegraphics[width=1\linewidth]{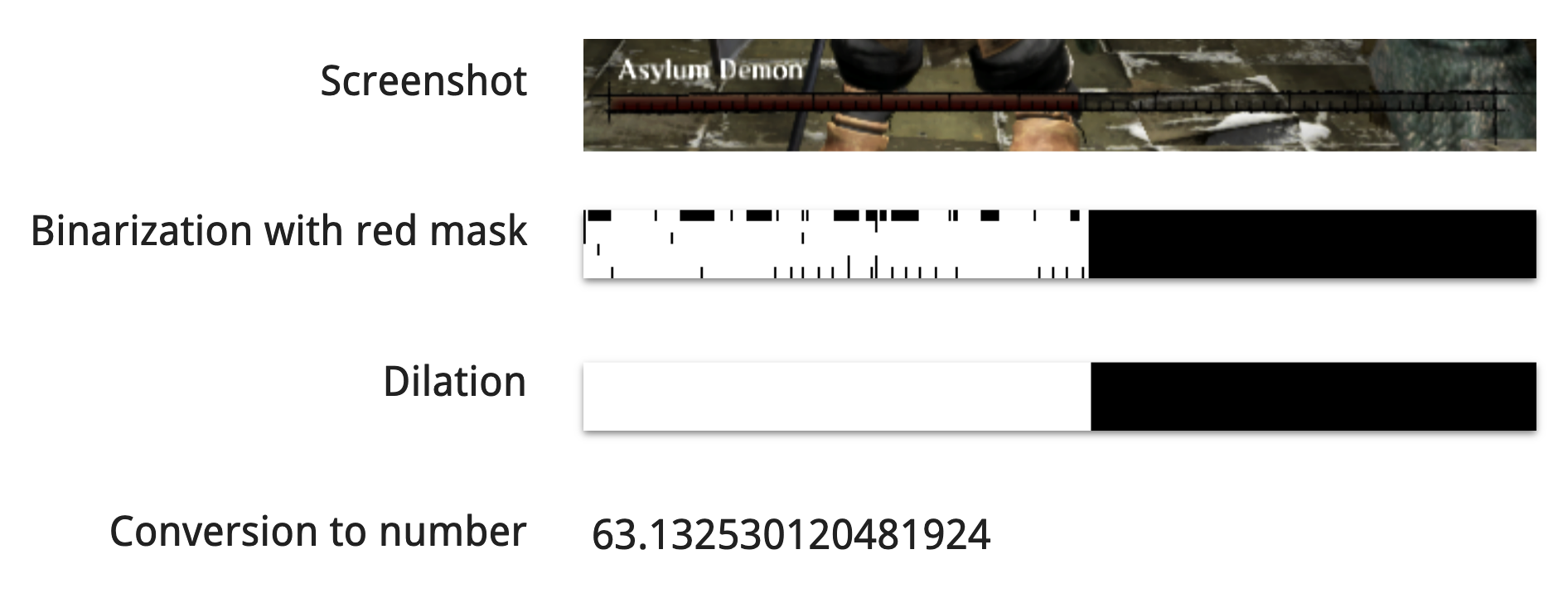}
    \caption{Health serves as a way to assess an agent's performance without explicitly rewarding a specific sequence of actions or hardcoding behavior. Without direct access to numerical data, alternative methods must be used to quantify visual information. Through DSAPI, the locations of the boss and player health bars are identified and cropped to match their size. A red mask is then applied to isolate the health bar from the background, followed by a dilation layer to refine the mask, after which the pixels are counted to get the value percentage.}
    \label{fig:fitnessfunction}
\end{figure}

\subsection{Experimental Design and Neural Network Configuration}

NEAT autonomously evolves neural network topology and weights, while input and output layer definitions are determined manually based on experimental design decisions. The games resolution was set at the lowest allowable value of 800×600 pixels to manage computational load. Screenshots were subsequently reduced to 40×30 pixels (1200 pixels per color channel), resulting in 3600 input neurons for RGB color data. This input dimensionality was selected based on preliminary evaluations balancing visual information retention and computational efficiency.

\subsection{Evaluation Metrics}

Fitness scores are calculated continuously by monitoring player and boss health using DSAPI, which employs HSV color masking and pixel counting to translate visual data into numeric fitness values. Currently, performance evaluation focuses solely on immediate combat outcomes. Potential alternative metrics for future evaluation could include accumulated Souls or environmental navigation efficiency.

Agents are evaluated based on the previously described fitness function, which measures offensive effectiveness, and survivability metrics. Performance data collected across generations allows for analysis of evolutionary progression and adaptation of network structures and combat strategies. The described experimental design establishes a controlled environment for assessing the application of vision-based NEAT in complex visual gameplay tasks.

\section{Results}

Our experimental evaluation involved training NEAT-evolved agents using only raw color visual inputs scaled to a resolution of 40×30 pixels. Under these conditions, evolved agents consistently reached mean fitness scores ranging from 70 to 80 (see Figure 3), where a maximum fitness of 200 corresponds to defeating the boss with full player health remaining. Agents achieved a maximum success rate of approximately 35\% over multiple generations, as shown in Figure 4, indicating a substantial win rate in a challenging game scenario.

Analysis of evolved neural networks indicated the emergence of specific strategies; for example, agents frequently adopted reactive healing behaviors, using the Estus Flask after sustaining damage and resuming aggressive actions upon recovery. In another instance, agents would focus on moving and rolling but striking when the opportunity presented itself.

\begin{figure}
    \centering
    \includegraphics[width=1\linewidth]{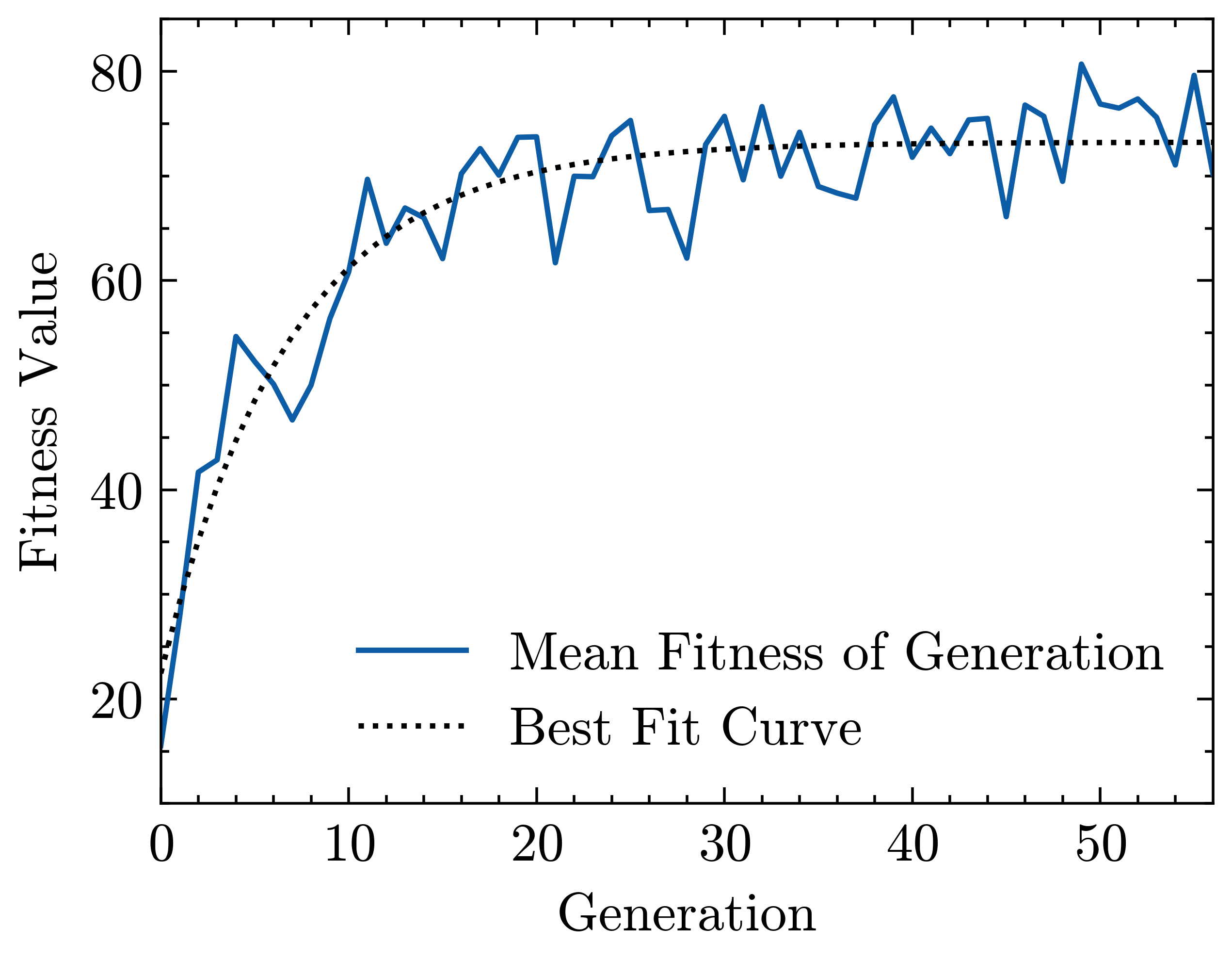}
    \caption{The mean fitness over generations. The solid blue line represents the mean fitness across all individuals per generation, while the dotted line depicts the best fit curve, illustrating the overall trend of fitness improvement across generations.}
    \label{fig:meanfitnessgraph}
\end{figure}

\begin{figure}
    \centering
    \includegraphics[width=1\linewidth]{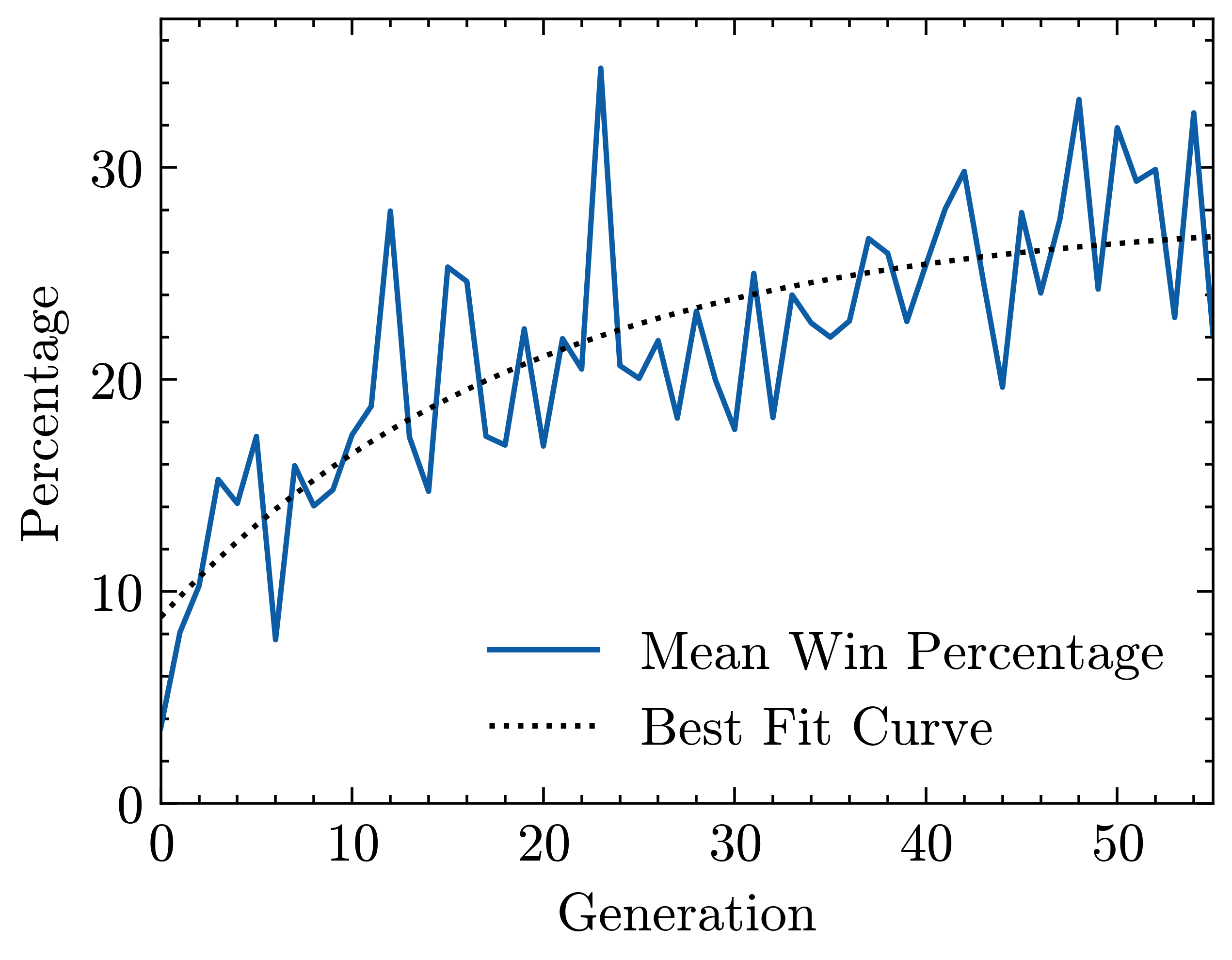}
    \caption{The mean win rate percentage over generations. The solid blue line represents the mean win rate across all individuals per generation, and the dotted line depicts the best fit curve. Note that individual agents may have a higher or lower win rate.}
    \label{fig:winpercentagegraph}
\end{figure}

\section{Conclusion}
This research demonstrates the applicability of vision-based neuroevolution, specifically NEAT, within the challenging context of Dark Souls. The experiments conducted against the Asylum Demon indicate that evolved agents can effectively learn combat behaviors purely from visual input, achieving notable levels of performance without explicit environmental information or traditional API access.

The use of DSAPI establishes a flexible, non-intrusive method for interacting with complex games, providing a replicable and scalable approach to AI research in environments lacking native programmatic interfaces. This approach also reduces potential ethical concerns related to game modification or code injection, offering an accessible pathway for extensive research within commercial gaming environments.

Furthermore, Dark Souls, due to its inherent complexity, challenging enemy behaviors, and nuanced combat mechanics, represents a compelling benchmark for vision-based AI research. Its difficulty makes it a valuable target for AI research, as successful strategies developed in this context may generalize to other demanding real-time decision-making scenarios. The success demonstrated in this initial study highlights the potential for significant improvements through advanced methods and increased computational resources.

\subsection{Future Work}
Future work includes exploring advanced processing of input images, incorporating additional gameplay actions, and extending evaluation scenarios to include environmental navigation and more complex enemy interactions. Further experiments can also investigate alternative neuroevolutionary methods and hybrid approaches combining reinforcement learning techniques to enhance agent performance and adaptability across diverse game scenarios.

\bibliographystyle{IEEEtran}
\bibliography{bibliography}

\end{document}